\documentclass[journal]{IEEEtran}

\ifCLASSINFOpdf
\else
\fi

\usepackage{graphicx}
\usepackage{amsmath,amssymb, bm} %
\usepackage{subfig}
\usepackage{color}
\usepackage[ruled]{algorithm2e}

\def\bal#1\eal{\begin{align}#1\end{align}} %

\newcommand{\pr}[1]{\left(#1\right)} %
\def\m{\mathbf}

\def\R{\mathbb{R}}
\def\ast{*}

\DeclareMathOperator*{\argmax}{arg\,max}
\DeclareMathOperator*{\argmin}{arg\,min}
\newcommand{\norm}[2]{\ensuremath{\left\|#1\right\|_{#2}}}

\newcommand {\bbmtx}{\begin{bmatrix}} %
\newcommand {\ebmtx}{\end{bmatrix}} %
\newcommand{\mean}[1]{\bar{#1}}
\usepackage[hidelinks,bookmarks=false]{hyperref}
\hypersetup{
  colorlinks   = true, %
  urlcolor     = blue, %
  linkcolor    = blue, %
  citecolor   = green %
}
\usepackage{multirow}
\usepackage[dvipsnames]{xcolor}

\begin{document}
\title{ Iterative Joint Image Demosaicking and Denoising using a Residual Denoising Network}

\author{Filippos~Kokkinos and Stamatios~Lefkimmiatis
\thanks{F. Kokkinos and S. Lefkimmiatis are with the Center for Computational and Data-Intensive Science and Engineering, Skolkovo Institute of Science and Technology.}%
}

\markboth{Journal of Transactions in Image Processing}
{Shell \MakeLowercase{\textit{et al.}}: Bare Demo of IEEEtran.cls for IEEE Journals}

\maketitle

\begin{abstract}
Modern digital cameras rely on the sequential execution of separate image processing steps to produce realistic images. The first two steps are usually related to denoising and demosaicking where the former aims to reduce noise from the sensor and the latter converts a series of light intensity readings to color images. Modern approaches try to jointly solve these problems, i.e. joint denoising-demosaicking which is an inherently ill-posed problem given that two-thirds of the intensity information is missing and the rest are perturbed by noise. While there are several machine learning systems that have been recently introduced to solve this problem, the majority of them relies on generic network architectures which do not explicitly take into account the physical image model. In this work we propose a novel algorithm which is inspired by powerful classical image regularization methods, large-scale optimization, and deep learning techniques. Consequently, our derived iterative optimization algorithm, which involves a trainable denoising network, has a transparent and clear interpretation compared to other black-box data driven approaches. Our extensive experimentation line demonstrates that our proposed method outperforms any previous approaches for both noisy and noise-free data across many different datasets. This improvement in reconstruction quality is attributed to the rigorous derivation of an iterative solution and the principled way we design our denoising network architecture, which as a result requires fewer trainable parameters than the current state-of-the-art solution and furthermore can be efficiently trained by using a significantly smaller number of training data than existing deep demosaicking networks.
\end{abstract}


\IEEEpeerreviewmaketitle

\section{Introduction}
Traditionally, high-resolution images from a digital camera are the end result of a processing pipeline that transforms light intensity readings to images. The image processing pipeline is typically modular, and the first two and most crucial steps involve image demosaicking and denoising. Due to the modular nature of the pipeline, demosaicking and denoising are dealt in a sequential manner where the ordering will either alter the light intensity readings from the sensor if denoising is applied first, or the initial demosaicking will introduce non-linearities in the noise statistics rending denoising an even harder problem. Moreover, both of these problems belong to the category of ill-posed problems while their joint treatment is very challenging since two-thirds of the underlying data are missing and the rest are perturbed by noise. Evidently, reconstruction errors during this early stage of the camera pipeline will produce unsatisfying final results.

Since demosaicking is an essential step of the camera pipeline, it has been extensively studied. For a complete survey of recent approaches, we refer to~\cite{li.2008}. One of the main drawbacks of the currently introduced methods that deal with the demosaicking problem, is that they assume a specific Bayer pattern\cite{li.2008,zhang2011color,duran2014self,buades2009self,heide2014flexisp,chang2015color,tan.2018}. This is a rather strong assumption and limits their applicability since there are many cameras available in the market that employ different Color Filter Array (CFA) patterns, for example, Fuji sensors. Therefore, demosaicking methods that are agile and able to generalize to different CFA patterns are preferred.

Typical methods that work for any CFA are the nearest neighbor and bilinear interpolation of the neighboring values for a given pixel for each channel. 
The problem with these approaches is the produced zippering artifacts which occur along high-frequency signal changes, e.g., edges and textures. Other classic approaches make use of the self-similarity and redundancy properties of natural images \cite{buades2009self,zhang2011color,duran2014self,chang2015color} to reconstruct the image, but they require an excessive amount of computation time and thus reducing their applicability to low-resource devices. Another successful class consists of methods that act upon the frequency domain, where any Bayer CFA can be represented as the combination of a luminance component at baseband and two modulated components~\cite{alleysson.2005}.

During recent years, research is directed towards learning based approaches, although a common problem with the design of learning based demosaicking algorithms is the lack of ground-truth images. In many approaches such as those in~\cite{sun.2013,he2012self, tan.2018} the authors used already processed images as references that are simulated mosaicked again, i.e. they apply a mosaick mask on the already demosaicked images, therefore obtaining non-realistic pairs for tuning trainable methods. Under this training strategy, the main issue is that demosaicking artifacts on the training data will hinder the performance and the overall quality of the reconstruction.  In a recent work Khasabi et al. \cite{khasabi2014} proposed a way to produce a dataset with realistic reference images allowing for the design of machine learning demosaicking algorithms. In their work, they thoroughly explained a methodology to create a demosaick dataset which is on par with reality.  We use the produced Microsoft Demosaicking dataset \cite{khasabi2014} to train, evaluate and compare our system. The reason is that the contained images have to be demosaicked in the linear RGB (linRGB) color space of the camera before being transformed via color transformation and gamma correction into standard RGB (sRGB) space that common consumer display devices use.  Furthermore, two popular CFA patterns are contained into the dataset, namely the Bayer and Fuji X Trans, which permits the development and evaluation of methods that can deal with different CFA patterns.

The effectiveness of neural networks for image demosaicking has been studied for over a decade. In earlier works \cite{Kapah.2000, Go.2000} feed forward neural networks were used on par with dictionary methods in order to obtain adaptive solutions for image demosaicking, while in~\cite{Wang.2014} small patches were used to train a multi-layer neural network minimizing an error function. In a recent work, Gharbi et. al.~\cite{Gharbi:2016:DJD:2980179.2982399} exploit the advantages in the field of deep learning to create a deep Convolutional Neural Network (CNN) that is able to demosaick images, and a lot of effort was put by the authors to create a new large demosaicking dataset, namely the MIT Demosaicking Dataset which consists of 2.6 million patches of images. Consequently, new CNN approaches were developed extending the usage of CNNs in the field. In Tan et al.~\cite{tan.2018} an ensemble of CNNs was developed which contained different models trained to demosaick patches with specific attributes, for example, textures and smooth areas, while Henz et al.~\cite{Henz.2018} constructed a convolutional autoencoder which was able to jointly design CFA and demosaick, therefore they obtained a CFA that out-performed the common Bayer CFA for image reconstruction purposes.

Apart from the demosaicking problem, another problem that requires special attention is the elimination of noise arising from the sensor and which distorts the acquired raw data. Firstly, the sensor readings are corrupted with shot noise~\cite{foi.2008} which is the result of random variation of the detected photons. Second, electronic inefficiencies during reading and converting the electrical charge into a digital count exhibit another type of noise, namely read noise. While shot noise is generally attributed to following a Poisson distribution and the read noise a Gaussian distribution; under certain circumstances, both noises can be approximated by a random variable following a heteroscedastic Gaussian pdf~\cite{foi.2008}. Prior work from Kalevo and Rantanen~\cite{kalevo.2002}, analyzed whether denoising should occur before or after the demosaicking step. It was experimentally confirmed that denoising is preferably done before demosaicking, however, Farsiu et al.~\cite{Farsiu.2006} formulates the solution of a joint estimation process in one step and demonstrates the superiority to an approach that breaks the problem into individual step. Later on, many researchers~\cite{menon.2009,zhang.2009,klatzer2016} validated experimentally the advantage of a joint estimation. Motivated by this long-known fact, we also pursue a joint approach for denoising and demosaicking of raw sensor data. 

In our previous work~\cite{kokkinos.2018}, we proposed an iterative neural network for solving the joint denoising-demosaicking problem that yielded state-of-the-art results on various datasets, both real and synthetic, without the need of millions of training images. In this work, we extend our previous work by dealing with some of its shortcomings while we further improve the reconstruction quality of our results. In detail, the extensions of our earlier work are: \textbf{(1)} A novel training strategy that allows us to train our network for an arbitrary number of iterations. In our previous work and other similar proposed iterative methods for different image restoration problems, the number of iterations is kept to a low number, usually no more than 10, due to memory restrictions. To the best of our knowledge, this is the first work that introduces a strategy that surpasses previous computational limitations and allows the training of iterative networks for an arbitrary number of iterations. The provided quantitative results across various datasets showcase the importance of our training strategy. \textbf{(2)} A modified version of our network that can better adapt to the noise level distorting the input image since it also accepts as a second argument the standard deviation of the noise. In our previous work~\cite{kokkinos.2018}, we handled the noise identically for every image, but as we will present in the sequel, a noise adaptive approach leads to considerable better image reconstruction quality. \textbf{(3)} Finally, we perform an extensive line of experimentation and ablation study to identify the performance of the proposed method with different configurations such as the number of iterations and the effect of different initialization schemes to the overall quality of the reconstruction.

\section{Problem Formulation}
To solve the joint demosaicking-denoising problem, one of the most frequently used approaches in the literature relies on the following linear observation model
\begin{equation}
\label{eq:linearmodel}
\m y = \m M \m x+ \m n,
\end{equation}
which relates the observed sensor raw data, $\m y \in \R^N$, and the underlying image $\m x \in \R^N$ that we aim to restore. Both $\m x$ and $\m y$ correspond to the vectorized forms of the images assuming that they have been raster scanned using a lexicographical order. Under this notation, $\m M\in \R^{N \times N}$ is the degradation matrix that models the spatial response of the imaging device, and in particular the CFA pattern. According to this, $\m M$ corresponds to a square diagonal binary matrix where the zero elements in its diagonal indicate the spatial and channel locations in the image where color information is missing. Apart from the missing color values, the image measurements are also perturbed by noise which hereafter, we will assume that is an i.i.d Gaussian noise $\m n \sim \mathcal{N}(0,\,\sigma^2)$. Note, that this is a rather simplified assumption about the noise statistics distorting the measurements. Nevertheless, our derived data-driven algorithm will be trained and evaluated on images that are distorted by noise, which follows statistics that better approximate real noisy conditions~\cite{foi.2008}.

Recovering $\m x$ from the measurements $\m y$ belongs to the broad class of linear inverse problems. For the problem under study, the operator $\m M$ is clearly singular, i.e. not invertible. This fact combined with the presence of noise perturbing the measurements leads to an ill-posed problem where a unique solution does not exist. One popular way to deal with this, is to adopt a Bayesian approach and seek for the Maximum A Posteriori (MAP) estimator
\begin{equation}
\label{eq:map}
\begin{aligned}
{\m x}^\star {} & =  \argmax_{\m x} \log(p(\m x|\m y)) \\
            & =  \argmax_{\m x} \log(p(\m y| \m x)) + \log(p(\m x)),
\end{aligned}
\end{equation}
where $\log(p(\m y| \m x))$ represents the log-likelihood of the observation $\m y$ and $\log(p( \m x))$ represents the log-prior of $\m{x}$. Problem~\eqref{eq:map} can be equivalently re-casted as the minimization problem 
\begin{equation}
\label{eq:var}
{\m x}^\star = \argmin_{\m x} \frac{1}{2\sigma^2} \norm{\m y-\m M \m x}{2}^2 + \phi(\m x),
\end{equation}
where the first term corresponds to the negative log-likelihood (assuming i.i.d Gaussian noise of variance $\sigma ^2$) and the second term corresponds to the negative log-prior. According to the above, the restoration of the underlying image $\m x$, boils down to computing the minimizer of the objective function in Eq.~\eqref{eq:var}, which  consists of two terms. This problem formulation has also direct links to variational methods where the first term can be interpreted as the data-fidelity that quantifies the proximity of the solution to the observation and the second term can be seen as the regularizer, whose role is to promotes solutions that satisfy certain favorable image properties. 

In general, the minimization of the objective function 
\begin{equation}
\label{eq:variational}
Q(\m x)= \frac{1}{2\sigma^2}\norm{\m y-\m M \m x}{2}^2 + \phi(\m x)
\end{equation}
is far from a trivial task, because the solution cannot simply be obtained by solving a set of linear equations. From the above, it is now clear that there are two important challenges that need to be dealt with before we are in the position of deriving a satisfactory solution for our problem. The first one is to come up with an algorithm that can efficiently minimize $Q\pr{\m x}$, while the second one is to select an appropriate form for $\phi\pr{\m x}$, which will constrain the set of admissible solutions by promoting only those that exhibit the desired properties. 

In Section~\ref{sec:MM}, we focus on the first challenge, while in Section~\ref{sec:ResDNet} we discuss how it is possible to avoid making any explicit decisions for the regularizer (or equivalently the negative log-prior) by following a machine learning approach. As we will describe in detail, the proposed strategy will allow us to efficiently regularize our solution but without the need to explicitly learn the form of the regularizer $\phi\pr{\m x}$. 

\section{Majorization Minimization Method}\label{sec:MM}
One of the main difficulties in the minimization of the objective function in Eq.~\eqref{eq:variational} is the coupling that exists between the singular degradation operator, $\m M$, and the latent image $\m x$. To circumvent this difficulty, there are several optimization strategies available that we could use, with potential candidates being splitting variables techniques such as the Alternating Direction Method of Multipliers~\cite{boyd.2011} and the Split Bregman approach~\cite{goldstein2009split}. However, one difficulty that arises by using such methods is that they involve additional parameters that need to be tuned so that a satisfactory convergence speed to the solution is achieved. Unfortunately, there is not a simple and straightforward way to choose these parameters. For this reason, in this work, we will instead pursue a majorization-minimization (MM) approach~\cite{hunter2004tutorial,figueiredo2007majorization, lefkimmiatis.2013}, which does not pose such a requirement. Under this framework, as we will describe in detail, instead of obtaining the solution by minimizing \eqref{eq:variational}, we compute it iteratively via the successive minimization of surrogate functions. The surrogate functions provide an upper bound of the initial objective function \cite{hunter2004tutorial}, and they are simpler to deal with than the original objective function. 

Specifically, in the majorization-minimization (MM) framework, an iterative algorithm for solving the minimization problem 
\begin{equation}
{\m x}^* = \argmin_x Q\pr{\m x}
\end{equation}
takes the form~\cite{hunter2004tutorial}
\begin{equation} \label{eq:mm_iter}
\m x^{(t+1)} = \argmin_x \tilde{Q}(\m x;{\m x}^{(t)}),
\end{equation}
where $\tilde{Q}(\m x;{\m x}^{(t)})$ is the majorizer of the function $Q(\m x)$ at a fixed point ${\m x}^{(t)}$, satisfying the two conditions
\begin{equation}
\label{eq:prop_one}
\tilde{Q}(\m x;\m x^{(t)}) > Q(\m x), \forall \m x\ne \m x^{(t)}\quad \mbox{and}\quad 
\tilde{Q}(\m x^{(t)};\m x^{(t)}) = Q(\m x^{(t)}).
\end{equation}

Here, the underlying idea is that instead of minimizing the actual objective function ${Q}(\m x)$, we first upper-bound it by a suitable majorizer $\tilde{Q}(\m x;{\m x}^{(t)})$, and then minimize this majorizing function to produce the next iterate $\m x^{(t+1)}$. Given the properties of the majorizer, iteratively minimizing $\tilde{Q}(\cdot;{\m x}^{(t)})$ also decreases the objective function $Q(\cdot)$. In fact, it is not even required that the surrogate function in each iteration is minimized. It is sufficient to find a $\m x^{(t+1)}$ that only decreases it. 

To derive a majorizer for $Q\pr{\m x}$ we opt for a majorizer of the data-fidelity term (negative log-likelihood). In particular, we consider the following majorizer
\begin{equation}
\label{eq:majorizer}
\tilde{d}(\m x,{\m x}_0) = \frac{1}{2\sigma^2}\norm{\m y-\m M \m x}{2}^2 + d(\m x,\m x_0),
\end{equation}
where $d(\m x, \m x_0) = \frac{1}{2\sigma^2}(\m x-\m x_0)^T [ \alpha \m I- \m M^T \m M ](\m x- \m x_0)$ is a function that measures the distance between $\m x$ and $\m x_0$. Since $\m M$ is a binary diagonal matrix, it is an idempotent matrix, that is $\m M^{T} \m M= \m M$, and thus $d(\m x,\m x_0) = \frac{1}{2\sigma^2}(\m x- \m x_0)^T [ \alpha \m I- \m M ](\m x- \m x_0)$. According to the conditions in~\eqref{eq:prop_one}, in order $\tilde{d}(\m x, \m x_0)$ to be a valid majorizer,  we need to ensure that $d(\m x, \m x_0) \ge 0, \forall \m x$ with equality iff $\m x= \m x_0$. This suggests that $\alpha \m I- \m M$ must be a positive definite matrix, which only holds when $\alpha > \norm{\m M}{2} = 1$, i.e $\alpha$ is bigger than the maximum eigenvalue of the binary diagonal matrix $ \m M$. { Based on the above, by replacing the data-fidelity term in Eq.~\eqref{eq:variational} with the respective majorizer in Eq.~\eqref{eq:majorizer} and performing the necessary algebraic calculations, the upper-bounded version is finally written as

\begin{equation}
\label{eq:upperboundeq}
\tilde{Q}(\m x,\m x_0) = \frac{1}{2(\sigma/\sqrt{\alpha})^2} \norm{\m x- \m z}{2}^2 + \phi(\m x) + c,
\end{equation}
where $c = \frac{1}{2\sigma^2}(\norm{\m y}{2}^2 + \m x_0^T(\alpha \m I - \m M) \m x_0 - \alpha \norm{\m z}{2}^2) $ does not depend on x and thus is a constant, while $\m z = \m y + (\m I - \m M)\m x_0$. }

Notice that following this approach, we have managed to completely decouple the degradation operator $\m M$ from $\m x$ and we now need to deal with a simpler problem. In fact, the resulting surrogate function in Eq.~\eqref{eq:upperboundeq} can be interpreted as the objective function of a denoising problem, with $\m z$ being the noisy measurements that are corrupted by noise whose variance is equal to $\sigma^2/a$. This is a key observation that we will heavily rely on, in order to design our deep network architecture. In particular, now it is possible instead of selecting the form of $\phi\pr{\m x}$ and minimizing the surrogate function, to employ a denoising neural network in order to compute the solution of the current iteration.

Our idea is similar in nature to other recent image restoration approaches that have employed denoising networks as part of alternative iterative optimization strategies, such as RED~\cite{romano2017little}, $P^3$~\cite{Venkatakrishnan.2013}, SEM~\cite{klatzer2016} and IRCNN~\cite{zhang.2017}. However, an important difference is that in our case we completely avoid the introduction of any free parameter that needs to be tuned by the user. We also move one step further from the aforementioned approaches by finetuning our denoising neural network on available data which allows us to implicitly incorporate domain knowledge to our learned regularizer. This direction for solving the joint denoising-demosaicking problem is very appealing since by using training data we can implicitly learn the function $\phi\pr{\m x}$ and also minimize the corresponding surrogate function using a feed-forward network.

\section{Residual Denoising Network (ResDNet)}\label{sec:ResDNet}
\begin{figure*}[ht]
\centering
\includegraphics[width=16cm]{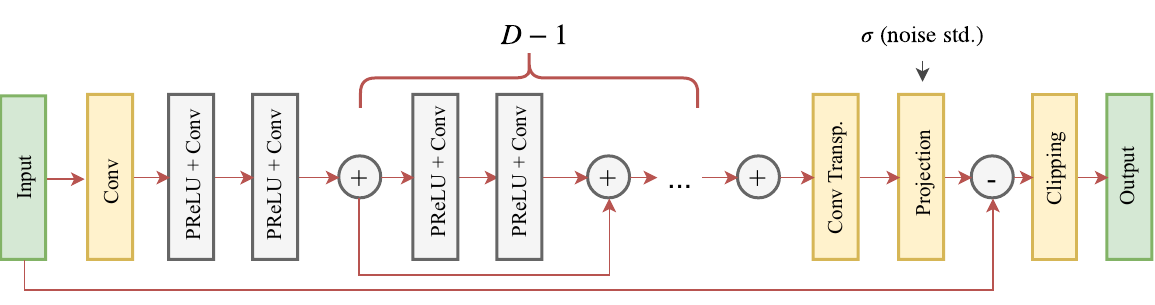}
\caption{Illustration of the employed residual denoiser; a core module of our framework. The deep learning based denoiser is used in every iteration to provide an approximate solution of the surrogate function.}
\label{fig:alg_n_denoiser}
\end{figure*}
Based on the discussion above, an important part of our approach is the design of a denoising network that will play the role of the solver for the surrogate function in Eq.~\eqref{eq:upperboundeq}. The architecture of the proposed network is depicted in Fig.~\ref{fig:alg_n_denoiser}. This is a residual network similar to DnCNN~\cite{DCNN}, where the output of the network is subtracted from its input. Therefore, the network itself acts as a noise estimator and its task is to estimate the noise realization that distorts the input. Such network architectures have been shown to lead to better restoration results than alternative approaches~\cite{DCNN,lefkimmiatis.2017}. One distinctive difference between our network and DnCNN, which also makes our network suitable to be used as a part of the MM-approach, is that it accepts two inputs, namely the distorted input and the variance of the noise. This way, as we will demonstrate in the sequel, we are able to learn a single set of parameters for our network and to employ the same network to inputs that are distorted by a wide range of noise levels. While the blind version of DnCNN can also work for different noise levels, as opposed to our network it features an internal mechanism to estimate the noise variance. However, when the noise statistics deviate significantly from the training conditions, such a mechanism can fail and thus DnCNN can lead to poor denoising results. In fact, due to this reason in~\cite{zhang.2017}, where more general restoration problems than denoising have been studied, the authors of DnCNN use a non-blind variant of their network as a part of their proposed restoration approach. However, training a deep network that requires a large number of parameters to be learned for each noise level can be rather impractical, especially in cases where one would like to employ such networks on devices with limited storage capacities. In our case, inspired by the recent work in~\cite{lefkimmiatis.2017} we circumvent this limitation by explicitly providing as input to our network the noise variance, which is then used to assist the network so as to provide an accurate estimate of the noise distorting the input. Note that there are several techniques available in the literature that can provide an estimate of the noise variance, such as those described in~\cite{Foi2009,Liu2013}, and thus this requirement does not pose any significant challenges in our approach.

A ResDNet with depth $D$, consists of five fundamental blocks. The first block is a convolutional layer with 64 filters whose kernel size is $5\times 5$. The second one is a non-linear block that consists of a parametrized rectified linear unit activation function (PReLU), followed by a convolutional layer with 64 filters of $3\times 3$ kernels. The PReLU function is defined as $\text{PReLU}(\m x) = \max(0, \m x) + \bm \kappa * \min(0,\m x)$ where $\bm \kappa$ is a vector whose size is equal to the number of input channels. In our network we use $D*2$ distinct non-linear blocks which we connect via a shortcut connection every second block in a similar manner to~\cite{he.2015} as shown in Fig.~\ref{fig:alg_n_denoiser}. Next, the output of the non-linear stage is processed by a transposed convolution layer which reduces the number of channels from 64 to 3 and has a kernel size of $5\times 5$. Then, it follows a projection layer~\cite{lefkimmiatis.2017} which accepts as an additional input the noise variance and whose role is to normalize the noise realization estimate so that it will have the correct variance before this is subtracted from the input of the network. Finally, the result is clipped so that the intensities of the output lie in the range $[0, 255]$. This last layer enforces our prior knowledge about the expected range of valid pixel intensities.

Regarding implementation details, before each convolutional layer, the input is padded to make sure that each feature map has the same spatial size as the input image. However, unlike the common approach followed in most of the deep learning systems for computer vision applications, we use reflexive padding than zero padding. Another important difference to other networks used for image restoration tasks~\cite{DCNN,zhang.2017} is that we do not use batch normalization after convolutions. Instead, we use the parametric convolution representation that has been proposed in~\cite{lefkimmiatis.2017} and which is motivated by image regularization related arguments. 

In particular, if $\m v\in \R^L$ represents the weights of a filter in a convolutional layer, these are parametrized as
\bal
\m v = s\frac{\pr{\m u -\mean{\m u}}}{\text{ }\norm{\m u -\mean{\m u}}{2}},
\eal
where $s$ is a scalar trainable parameter, $\m u\in \R^L$ and $\mean{\m u}$ denotes the mean value of $\m u$. In other words, we are learning zero-mean valued filters whose $\ell_2$-norm is equal to $s$.

Furthermore, the projection layer, which is used just before the subtraction operation with the network input, corresponds to the following $\ell_2$ orthogonal projection as in~\cite{lefkimmiatis.2017}
\begin{equation}
\mathcal{P_{\mathcal{C}}}\pr{\m y} =\varepsilon \frac{\m y}{\max(\norm{\m y}{2},\varepsilon)},
\end{equation}
where $\varepsilon = e^\gamma\sigma\sqrt{N-1}$, $N$ is the total number of pixels in the image (including the color channels), $\sigma$ is the standard deviation of the noise distorting the input, and $\gamma$ is a scalar trainable parameter. As we mentioned earlier, the goal of this layer is to normalize the noise realization estimate so that it has the desired variance before it is subtracted from the network input. 

\section{Demosaicking Network Architecture}
\begin{figure}[t]
\hspace{-0.8cm}
\includegraphics[width=10.5cm]{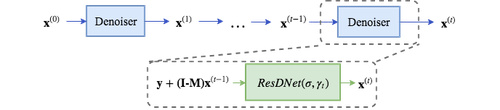}
\caption{A graphical representation of the proposed iterative neural network. We have omitted the extrapolation steps for clarity.}
\label{fig:alg}
\end{figure}

The overall architecture of our approach is based upon the MM framework, presented in Section~\ref{sec:MM}, and the proposed denoising network. As discussed, the MM is an iterative algorithm Eq.~\eqref{eq:mm_iter} where the minimization of the majorizer in Eq.~\eqref{eq:upperboundeq} can be interpreted as a denoising problem. One way to design the demosaicking approach would be to unroll the MM algorithm as $K$ discrete steps and then for each step use a different denoising network to retrieve the solution of Eq.~\eqref{eq:upperboundeq}. However, this approach can have two distinct drawbacks which will hinder its performance. The first one is that the usage of a different denoising neural network for each step like in \cite{zhang.2017}, demands a high overall number of parameters, which is equal to $K$ times the parameters of the employed denoiser, making the demosaicking network impractical for any real applications. Simultaneously, the high overall number of network parameters would require a significantly higher number of training data and training time. To override these drawbacks, we opt to use our ResDNet denoiser, which can be applied to a wide range of noise levels, for all $K$ steps of our demosaick network, using the same set of parameters. By sharing the parameters of our denoiser across all the $K$ steps, the overall demosaicking approach maintains a low number of parameters and requires only a few hundred of images to train.

The second drawback of the MM framework as described in Section~\ref{sec:MM} is the slow convergence \cite{fista.2009} that it can exhibit. Beck and Teboulle \cite{fista.2009} introduced an accelerated version of this MM approach which combines the solutions of two consecutive steps with a certain extrapolation weight that is different for every step. In this work, we adopt a similar strategy which we describe in Algorithm~\ref{alg:the_alg}. Furthermore, in our approach, we go one step further and instead of using the values originally suggested in~\cite{fista.2009} for the weights $\bm w \in \mathbb{R}^K$, we treat them as trainable parameters and learn them directly from the data. These weights are initialized with $w_i = \frac{i-1}{i+2}\text{,} \forall 1 \le i \le K$. The underlying reasoning of finetuning the extrapolation weights lies in \cite{li.lin.2015} where the authors claim that some extrapolation weights may hinder the convergence, so we opt to fit the weights upon available data and derive a data-driven extrapolation suitable for our problem.

The convergence of our framework can be further sped up by employing a continuation strategy \cite{Lin2015} where the main idea is to solve the problem in Eq.~\eqref{eq:upperboundeq} with a large value of $\sigma$  and then gradually decrease it until the target value is reached.  Our approach is able to make use of the continuation strategy due to the design of our ResDNet denoiser, which accepts as additional arguments the noise variance that is fed to the learnable projection layer, and the ability to denoise images for various noise levels. In detail, we initialize the trainable parameter of the projection layer $\bm \gamma \in \mathbb{R}^K$ with values spaced evenly on a log scale from $\gamma_{max}$ to $\gamma_{min}$ and later on the vector $\bm \gamma$ is further finetuned on the training dataset via back-propagation. 

In summary, our overall demosaicking network is described in  Algorithm~\ref{alg:the_alg} where the set of trainable parameters $\theta$ consists of the parameters of the ResDNet denoiser, the extrapolation weights $\bm w$ and the projection parameters $\bm \gamma$. All of the aforementioned parameters are initialized as described in the current section and Section~\ref{sec:ResDNet} and are trained on specific demosaick datasets.

\begin{algorithm}[t]
 \SetAlgoCaptionSeparator{\unskip:}
 \SetKwInOut{Input}{Input}
 \Input{$\m M$: CFA, $\m y$: input, $K$: iterations, $\bm w\in\R^K$: extrapolation weights, $\sigma$: estimated noise, $\bm \gamma\in\R^K$: projection parameters}%
 $\m x^{(0)}= \m 0$\;
 Initialize $\m x^{(1)}$ using $\m y$\;
  \For{$i\gets1$ \KwTo $K$}{
    $ \m u = \m x ^{(i)} + \bm w_i (\m x^{(i)} - \m x^{(i-1)}) $\;
    $\m x^{(i+1)} = \text{ResDNet}((\m I-\m M) \m u + \m y , \sigma, \bm \gamma_i )$\; 
 }
 \caption{The proposed demosaicking network described as an iterative process. The ResDNet parameters are shared across all iterations.}
  \label{alg:the_alg}
\end{algorithm}

Finally, while our demosaick network shares a similar philosophy with methods such as RED~\cite{romano2017little}, $P^3$~\cite{Venkatakrishnan.2013} and IRCNN~\cite{zhang.2017}, it exhibits some important and distinct differences. In particular, the strategies as mentioned above make use of certain optimization schemes to decompose their original problem into sub-problems that are solvable by a denoiser. For example, the authors of $P^3$~\cite{Venkatakrishnan.2013} decompose the original problem Eq.~\eqref{eq:linearmodel} relying on the  Alternating Direction Method of Multipliers (ADMM)~\cite{boyd.2011} and solve a linear system of equations and a denoising problem instead. The authors of RED~\cite{romano2017little} go one step further and employ a regularizer that involves an image-adaptive Laplacian, which in turn allows the use of several classical and machine-learning based denoisers to serve as sub-solvers. Both approaches are similar to ours; however, their formulation involves a tunable variable $\lambda$ that weights the participation of the regularizer on the overall optimization procedure. Thus, in order to obtain an accurate reconstruction in a timely manner, the user must manually tune the variable $\lambda$ which is tractable but not a trivial task. On the other hand, our method does not involve any tunable variables. 

Furthermore, the approaches $P^3$, RED  and IRCNN are based upon static denoisers like Non Local Means~\cite{buades2005non}, BM3D~\cite{dabov2007image} and DCNN~\cite{DCNN}, meanwhile we opt to use a universal denoiser, like ResDNet, that can be further trained on any available training data. In a similar fashion, the Deep Mean-Shift Priors~\cite{bigdeli2017deep} approach employs a denoising autoencoder to approximate the gradient of the learned prior which is then used in the minimization of a range of different objectives via gradient descent with momentum. However, a distinct difference of this method with our approach is the fact that our learned regularizer is task-specific and does not depend on tunable parameters. Finally, our approach goes one step further, and we use a trainable version of an iterative optimization strategy for the task of the joint denoising-demosaicking in the form of a feed-forward neural network.

\section{Network Training}

\subsection{Joint Denoising and Demosaicking} \label{sec:mmnet_train}
Since Eq.~\eqref{eq:upperboundeq} is, as already discussed, a denoising step, we pre-train our ResDNet denoiser on the simple case where $\m M=\m I$; casting our problem an Additive White Gaussian(AWGN) denoising task. We found experimentally that pre-training the ResDNet vastly reduces the necessary training time because it is a proper initialization for the joint denoising and demosaicking task. In detail, the denoising network ResDNet that we use as part of our iterative approach is pre-trained on the Berkeley segmentation dataset (BSDS) \cite{937655}, which consists of 500 color images. These images were split into two sets, 400 were used to form a train set, and the rest 100 formed a validation set. All the images were randomly cropped into patches of size $180 \times 180$ pixels. The patches were perturbed with noise $\sigma \in [0,15]$ and the network was optimized to minimize the Mean Square Error. We set the network depth $D=5$, all weights are initialized as in He et al. \cite{he.delving} and the optimization is carried out using AMSGRAD \cite{j.2018on} which is a stochastic gradient descent algorithm that adapts the learning rate per parameter. The training procedure starts with an initial learning rate equal to $10^{-2}$.

The pre-trained denoiser used in the presented iterative approach is further trained end-to-end to minimize the averaged $L_1$ loss over a mini-batch of size $d$.

\begin{equation}
\label{eq:loss}
L(\theta) = \frac{1}{N}\sum_{i=1}^d \norm{\m y_i - f(\m x_i)}{1},
\end{equation}
where $\m y_i \in \mathbb{R}^{N}$ and $\m x_i \in \mathbb{R}^{N}$ are the rasterized ground-truth and input images, while $f\left(\m x_i\right)$ denotes the output of our proposed network. Due to the iterative nature of our framework, the network parameters are updated using the Back-propagation Through Time (BPTT) algorithm. Specifically, using the same denoiser for every iteration in Alg.~\ref{alg:the_alg} means that the same set of parameters is used for every iteration and thus we have to sum the parameter changes in the $K$ unfolded instances in order to train the network.  However, if the number of total iterations $K$ is high, then a number of prohibitive restrictions arise during training, for example, both $K$ and mini-batch size $d$ are upper-bounded from the GPU memory consumption. It is evident that we need to keep all intermediate results of the Alg.~\ref{alg:the_alg} in order to calculate the gradients which increases the memory requirements by a factor of $K$. Thankfully, there is a workaround to avoid such restrictions by using the Truncated Back-propagation Through Time (TBPTT)~\cite{robinson.2008} algorithm instead, which we explain in detail in Section~\ref{sec:TBPTT}. Using this solution, we are able to use an arbitrary number of total iterations $K$ for training and increase the batch size $d$ with the only trade-off of being the increase in the computation time. Furthermore, the optimization is carried again via the AMSGRAD optimizer and the training starts from a learning rate of $10^{-2}$ which we decrease it by a factor of 10 every 100 epochs. Finally, for every noise-free experiment we set $\gamma_{max}=15$ and $\gamma_{min}=0$, while for every other case the respective values are $\gamma_{max}=2$ and $\gamma_{min}=0$. %

\begin{figure}[t]
\centering
\includegraphics[width=9cm]{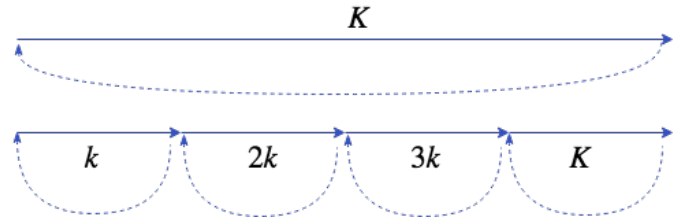}
\caption{Visualization of the differences between the BPTT and TBPTT algorithms, where inference is depicted as a solid line and back-propagation as a dashed line. By employing TBPTT, we unroll the network's $K$ iterations and split them into several stages, where each stage involves a smaller number of $k$ iterations. Then the network parameters are updated at the end of each stage, as opposed to the BPTT algorithm where the parameters are updated only once at the end of the $K$th iteration. In our implementation, we apply a weight to each loss function for all stages except to the last one to impose that the last iteration will produce the best result. }
\label{fig:TBPTT}
\end{figure}
\subsection{Training with TBPTT}
\label{sec:TBPTT}

\begin{figure}[!ht]
\centering
\includegraphics[width=9cm]{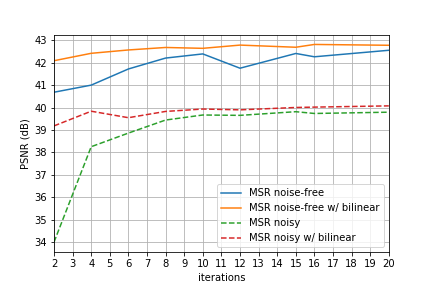}
\caption{By increasing the number of iterations, our deep network is capable of achieving increasingly better reconstruction with the same number of parameters. Also, it can be seen that given a large number of iterations; the method is capable of achieving virtually the same performance with and without proper initialization.}
\label{fig:psnr_vs_iter}
\end{figure}

As discussed, we would like to use an arbitrary amount of iterations $K$ during training, however, this is not always possible because $K$ is upper-bounded by the available GPU memory or RAM. Consequently, a higher number of iterations force us to use smaller batches casting the training of our framework slow. Therefore, to overcome this restriction we propose to train our framework with the TBPTT algorithm where the loop is unrolled into a small set of $k$ iterations (stages) out of $K$ and the back-propagation is performed sequentially on the unrolled stages as pictured in Fig.~\ref{fig:TBPTT}. Thankfully, TBPPT comes with no additional computational cost because an optimized implementation of TBPTT has the same computation complexity with the full BPTT. As presented in Fig.~\ref{fig:TBPTT} we compute the loss Eq.~\eqref{eq:loss} after $k$ iterations.  The calculation of multiple loss functions makes difficult the retrieval of the best reconstruction result during inference because this may occur in an earlier iteration rather than the last. To avoid this problem, we weight all intermediate loss function calculations by a factor of 0.5 and only the last calculation of the loss function remains unweighted. Our empirical training approach makes possible the retrieval of the best reconstruction at the last iteration.  

\begin{figure}[!t]
\centering
\includegraphics[width=9cm]{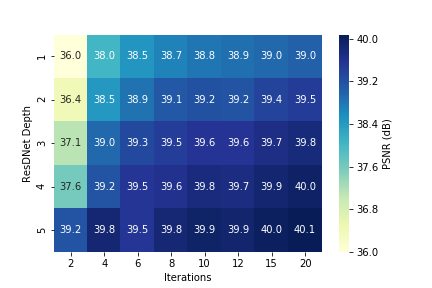}
\caption{Quantitative PSNR comparison between the depth of the denoiser and the number of iterations on the noisy MSR Dataset. The heat-map depicts the ability of our method to generalize by decreasing the number of parameters and simultaneously increasing the number of iterations.}
\label{fig:heatmap}
\end{figure}

In similar learnable iterative approaches such as those in~\cite{dong.2018,jian.zhang.2017} the authors argued that a few iterations between 4 to 6 are enough to obtain descent reconstruction quality. However, we experimentally found that while a few number of iterations might lead to adequate results aimed mostly on applications where computational time needs to be kept relatively low, the best results can be achieved with as many as 20 iterations or even more. Indeed, as presented in Figs.~\ref{fig:psnr_vs_iter} and~\ref{fig:heatmap} the Peak Signal-to-Noise Ratio (PSNR) is increasing till a certain number of iterations, and after that it stabilizes. Also, from Fig.~\ref{fig:psnr_vs_iter} it is clear that a proper initialization of the network's input allows the use of only a few iterations in order to retrieve comparable performance. Nevertheless, our method is capable of achieving the same performance even with an improper initialization if a higher number of iterations is used, making our approach very attractive for other inverse problems where no good initialization exist such as compressed sensing. Finally, in Fig.~\ref{fig:heatmap} the trade-off between the number of network parameters and the iterations is demonstrated. In particular, from these results we observe that methods which involve a denoising network of smaller depth and thus fewer trainable parameters can produce meaningful results if the number of iterations is high enough, while methods that depend on deeper denoising networks require only a small number of iterations in order to match the same performance.

\section{Experiments}
\label{sec:experiments}
We perform an extensive line of experimentation on multiple datasets and CFA patterns in order to evaluate and analyze our method. Our main metric used for comparisons among different methods in all reported experiments is the PSNR.

\subsection{Demosaicking Artificial Data}
At first, we compare our method against prior work on the pure demosaicking task with artificially created mosaicked data on the $sRGB$ color space. The artificial data are created using the standard datasets McMaster~\cite{zhang2011color}, Kodak~\cite{li.2008}\footnote{The Kodak dataset is re-sized to 512x768 following the methodology of evaluation described in~\cite{li.2008}.} and the newly created MIT dataset~\cite{Gharbi:2016:DJD:2980179.2982399}; all of which are 8-bit sRGB data and they are re-mosaicked, so the following experiments deviate from the standard camera pipeline used in practical scenarios. Beyond this, the aforementioned dataset is known to have flaws \cite{Gharbi:2016:DJD:2980179.2982399} and misrepresent the statistics of natural images \cite{levin.2012}. Due to the fact that these images are noise-free, we manually set $\sigma=1$ as the noise standard deviation for all images and no noise estimation takes place. The whole MIT Dataset is used for training for 10 epochs, and we set $K=1$ in order to speed up the training process; after that, the network is trained on a small random subset of 40.000 with $K=10$. In this case, the input is the mosaicked image, and no initialization takes place. The whole training procedure requires approximately four days. From the reported results in Table~\ref{tbl:general} we observe that our network achieves comparable performance for all different datasets with current state-of-the-art approaches, although, it requires only a fraction of the parameters that the other systems use. In more detail, our method for the demosaick-only scenario is using 380.356, while Gharbi et al.~\cite{Gharbi:2016:DJD:2980179.2982399} and Henz et al.~\cite{Henz.2018} make use of 559.776 and 1.204.780 parameters respectively.

\begin{table}[t]
\centering
\begin{tabular}{ccccc}
\hline \hline
\multicolumn{1}{l}{} & Kodak & McM & VDP & Moire \\ \hline
Bilinear & 32.9 & 32.5 & 25.2 & 27.6 \\
Zhang (NLM) \cite{zhang2011color} & 37.9 & 36.3 & 30.1 & 31.9 \\
Hirakawa \cite{Hirakawa.2005} & 36.5 & 33.9 & 30.0 & 32.1 \\
Getreuer \cite{getreuer.2011} & 38.1 & 36.1 & 30.8 & 32.5 \\
Heide \cite{heide2014flexisp} & 40.0 & 38.6 & 27.1 & 34.9 \\
Klatzer \cite{klatzer2016} & 35.3 & 30.8 & 28.0 & 30.3 \\
Gharbi \cite{Gharbi:2016:DJD:2980179.2982399} & 41.2 & 39.5 & 34.3 & 37.0 \\
Kokkinos \cite{kokkinos.2018} & 41.5 & \textbf{39.7} & \textbf{34.5} & 37.0 \\
Henz \cite{Henz.2018} & 41.9 & 39.5 & 34.3 & 36.3 \\
$\text{MMNet}_{10}$ (ours) & \textbf{42.0} & \textbf{39.7} & \textbf{34.5} & \textbf{37.1} \\ \hline
\end{tabular}
\caption{Comparison of our system to state-of-the-art techniques on the demosaick-only scenario on artificial data in terms of Peak signal-to-noise ratio (PSNR) performance. }
\label{tbl:general}
\end{table}

\subsection{Demosaicking Raw Data}
\begin{figure*}[!ht]
  \centering
  \subfloat[]{\includegraphics[width=0.25\textwidth, height=0.23\textwidth]{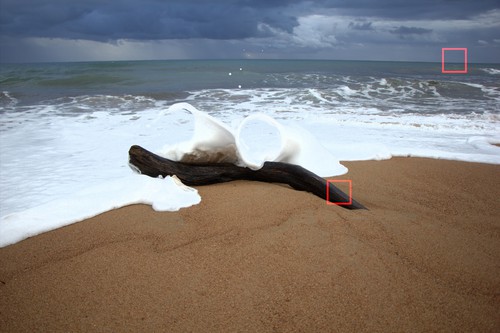}}
  \subfloat[]{\includegraphics[width=0.25\textwidth, height=0.23\textwidth]{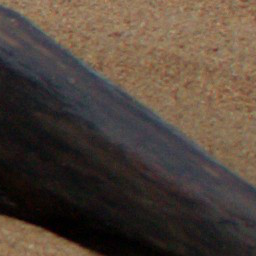}}
  \subfloat[]{\includegraphics[width=0.25\textwidth, height=0.23\textwidth]{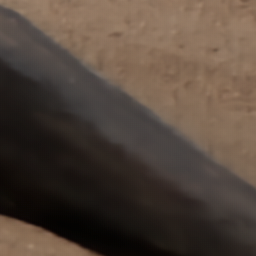}}
  \subfloat[]{\includegraphics[width=0.25\textwidth, height=0.23\textwidth]{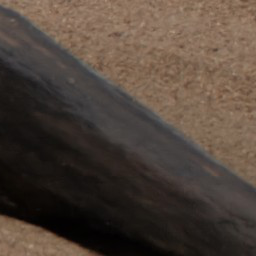}} \\
  \vspace{-0.04\textwidth}
  \hfill
  \subfloat[]{\includegraphics[width=0.25\textwidth, height=0.23\textwidth]{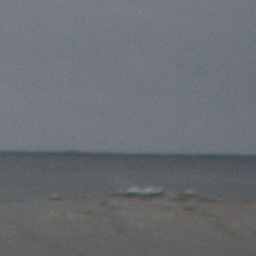}}
  \subfloat[]{\includegraphics[width=0.25\textwidth, height=0.23\textwidth]{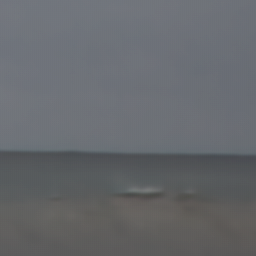}}
  \subfloat[]{\includegraphics[width=0.25\textwidth, height=0.23\textwidth]{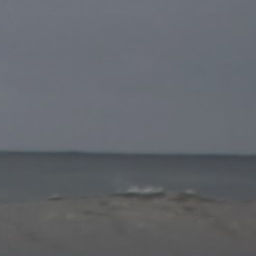}} \\
  \vspace{-0.04\textwidth}
  \subfloat[]{\includegraphics[width=0.25\textwidth, height=0.23\textwidth]{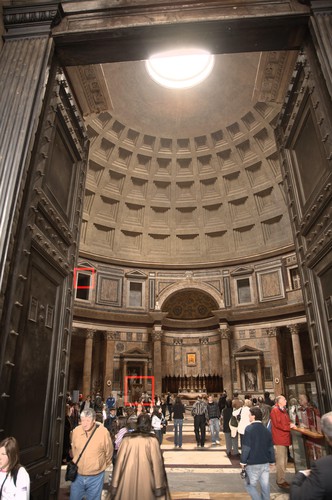}}
  \subfloat[]{\includegraphics[width=0.25\textwidth, height=0.23\textwidth]{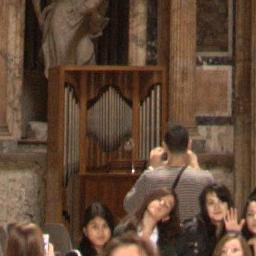}}
  \subfloat[]{\includegraphics[width=0.25\textwidth, height=0.23\textwidth]{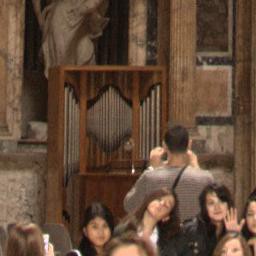}}
  \subfloat[]{\includegraphics[width=0.25\textwidth, height=0.23\textwidth]{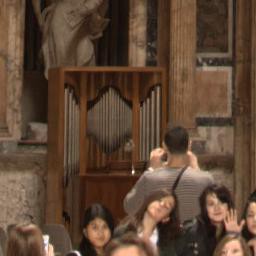}} \\
  \vspace{-0.04\textwidth}
  \hfill
  \subfloat[]{\includegraphics[width=0.25\textwidth, height=0.23\textwidth]{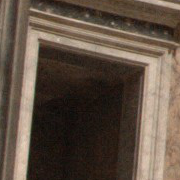}}
  \subfloat[]{\includegraphics[width=0.25\textwidth, height=0.23\textwidth]{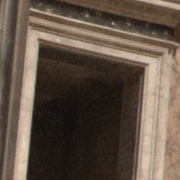}}
  \subfloat[]{\includegraphics[width=0.25\textwidth, height=0.23\textwidth]{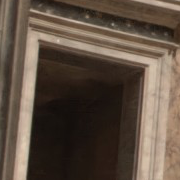}} \\
  \vspace{-0.04\textwidth}
  \subfloat[Ours (full)]{\includegraphics[width=0.25\textwidth, height=0.23\textwidth]{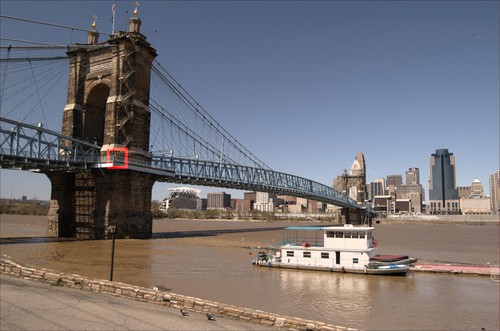}}
  \subfloat[Hirakawa \cite{Hirakawa.2005} (dcraw)]{\includegraphics[width=0.25\textwidth, height=0.23\textwidth]{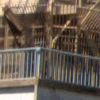}}
  \subfloat[Gharbi et al. \cite{Gharbi:2016:DJD:2980179.2982399}]{\includegraphics[width=0.25\textwidth, height=0.23\textwidth]{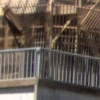}}
  \subfloat[ours]{\includegraphics[width=0.25\textwidth, height=0.23\textwidth]{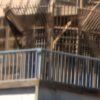}} \\
  \caption{Performance of common systems on real RAW images. Our method achieves a high-quality reconstruction without exhibiting any significant noise artifacts. Note that in the challenging scenario depicted in the last row, the alternative methods introduce moire and blocking artifacts on the rail of the bridge. Results are best viewed by zooming in on a computer screen.}
  \label{fig:showcase1}
\end{figure*}

As mentioned earlier, Khashabi et al.~\cite{khasabi2014} proposed that the evaluation of demosaicking and denoising should occur on raw RGB data because this testing pipeline is closely related to real digital imaging applications. MSR dataset contains exclusively linear data encoded in the standard imaging 16-bit representation. The dataset contains 200 images for training, which we augment with vertical and horizontal flips, 100 for validation purposes and a test set of 200 images. The same 200 training images are provided with noise perturbations using the affine noise model \cite{foi.2008} with unknown values for the hyper-parameters of the noise. The lack of these hyperparameters renders difficult the production of training data that follow the same noise statistics. This reason, forced Gharbi et al.~\cite{Gharbi:2016:DJD:2980179.2982399} to use the simplistic approach of Gaussian noise for training data creation which diverges greatly from practical applications and this fact is reflected from the inferior performance that their system achieves in the noisy case (38.6 dB). While we made the same simplistic assumption in Section~\ref{sec:MM}, in order to design the network architecture, our network after being trained on more realistic noise conditions, is capable of denoising successfully even the data dependent part of the affine noise models, as it can be seen in Fig.~\ref{fig:showcase1}. We estimate $\sigma$ for each raw image using the median absolute deviation of the wavelet detail coefficients as described in \cite{donoho.1994} and provide it to the network as additional input.

\begin{figure*}[!htbp]
  \centering
\subfloat{\includegraphics[width=0.2\textwidth]{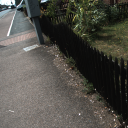}}
  \subfloat{\includegraphics[width=0.2\textwidth]{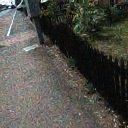}}
  \subfloat{\includegraphics[width=0.2\textwidth]{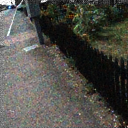}}
  \subfloat{\includegraphics[width=0.2\textwidth]{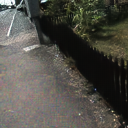}}
  \subfloat{\includegraphics[width=0.2\textwidth]{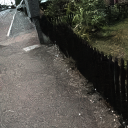}} \\
  \vspace{-0.015\textwidth}
  \subfloat{\includegraphics[width=0.2\textwidth]{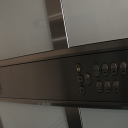}}
  \subfloat{\includegraphics[width=0.2\textwidth]{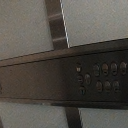}}
  \subfloat{\includegraphics[width=0.2\textwidth]{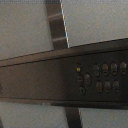}}
  \subfloat{\includegraphics[width=0.2\textwidth]{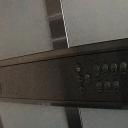}}
  \subfloat{\includegraphics[width=0.2\textwidth]{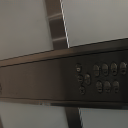}} \\ 
  \vspace{-0.015\textwidth}
  \subfloat{\includegraphics[width=0.2\textwidth]{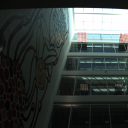}}
  \subfloat{\includegraphics[width=0.2\textwidth]{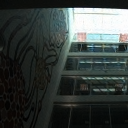}}
  \subfloat{\includegraphics[width=0.2\textwidth]{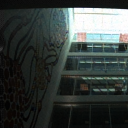}}
  \subfloat{\includegraphics[width=0.2\textwidth]{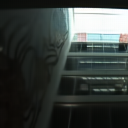}}
  \subfloat{\includegraphics[width=0.2\textwidth]{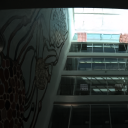}} \\
  \vspace{-0.015\textwidth}
  \subfloat{\includegraphics[width=0.2\textwidth]{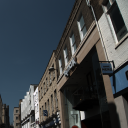}}
  \subfloat{\includegraphics[width=0.2\textwidth]{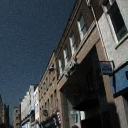}}
  \subfloat{\includegraphics[width=0.2\textwidth]{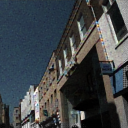}}
  \subfloat{\includegraphics[width=0.2\textwidth]{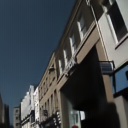}}
  \subfloat{\includegraphics[width=0.2\textwidth]{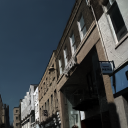}} \\
  \vspace{-0.015\textwidth}
  \subfloat{\includegraphics[width=0.2\textwidth]{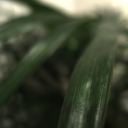}}
  \subfloat{\includegraphics[width=0.2\textwidth]{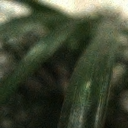}}
  \subfloat{\includegraphics[width=0.2\textwidth]{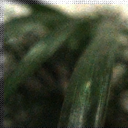}}
  \subfloat{\includegraphics[width=0.2\textwidth]{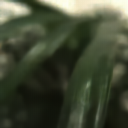}}
  \subfloat{\includegraphics[width=0.2\textwidth]{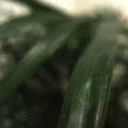}} \\
  \vspace{-0.015\textwidth}
  \subfloat[Reference]{\includegraphics[width=0.2\textwidth]{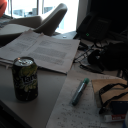}}
  \subfloat[Hirakawa (dcraw)\cite{Hirakawa.2005}]{\includegraphics[width=0.2\textwidth]{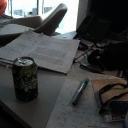}}
  \subfloat[Zhang(NLM)\cite{zhang2011color}]{\includegraphics[width=0.2\textwidth]{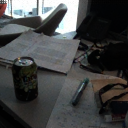}}
  \subfloat[Gharbi et al.\cite{Gharbi:2016:DJD:2980179.2982399}]{\includegraphics[width=0.2\textwidth]{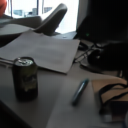}}
  \subfloat[ours]{\includegraphics[width=0.2\textwidth]{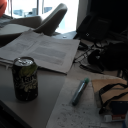}}
  \caption{Comparison of our network with other competing techniques on images from the noisy MSR Dataset. From these results it is clear that our method is capable of removing the noise while keeping fine details. On the contrary, the rest of the methods either fail to denoise or they oversmooth the images.}
  \label{fig:real_images}
\end{figure*}

Surprisingly, as show in Table~\ref{tbl:msr} our method is capable to achieve slightly better performance to Gharbi et al. in the noise-free scenario using only 200 training images and surpass previous approaches in the noisy scenario by a large margin of 1.3 dB if we employ $K=20$ iterations and 0.4 dB if we set $K=2$. This fact clearly demonstrates the capabilities of our method to generalize better even when trained on small datasets and the ability to trade off performance for computing time and vice versa. Apart from the quantitative results in Fig.~\ref{fig:real_images}, we showcase and compare the results of our method and other well known approaches on real RAW images acquired from the web. Clearly, our method leads to better and more visually pleasing results, since it is able to demosaick the RAW image restoring the fine details while effectively suppressing noise. {The training time for MSR dataset is linear to the number of iterations therefore for   $K=2$ training takes  approximately 1 hour and for $K=20$  nearly 11 hours are required. }
\begin{table}[!t]
\centering
\begin{tabular}{lcccc}
\hline \hline
\multicolumn{1}{l}{}                                        & \multicolumn{2}{c}{noisy}              & \multicolumn{2}{c}{noise-free}                   \\
\multicolumn{1}{l}{}                                        & linRGB               & sRGB                 & linRGB               & sRGB                 \\
Bilinear                                                    & -                    & -                    & 30.9                 & 24.9                 \\
Zhang(NLM) \cite{zhang2011color}                            & -                    & -                    & 38.4                 & 32.1                 \\
Hirakawa \cite{Hirakawa.2005}                               & -                    & -                      & 37.2                      & 31.3          \\
Getreuer \cite{getreuer.2011}                               & -                    & -                    & 39.4                 & 32.9                 \\
Heide \cite{heide2014flexisp}                               & -                    & -                    & 40.0                 & 33.8                 \\
Khasabi \cite{khasabi2014}                                  & 37.8                 & 31.5                 & 39.4                 & 32.6                 \\
Klatzer \cite{klatzer2016}                                  & 38.8                 & 32.6                 &  40.9        & 34.6                 \\
Gharbi \cite{Gharbi:2016:DJD:2980179.2982399}                                            & 38.6                 & 32.6                 & 42.7                 & 35.9                 \\
 Bigdeli \cite{bigdeli2017deep}                                            & 38.7                 & -                 & -                 & -                 \\

Kokkinos \cite{kokkinos.2018}                                            & 39.2                 & 33.3                 & 41.0                 & 34.6                 \\
$\text{MMNet}_{2}$  (ours)                                  & 39.2                 & 33.3                 &  42.1        & 35.6                \\
$\text{MMNet}_{20}$  (ours)                                 & \textbf{40.1}        & \textbf{34.2}        & \textbf{42.8}                 & \textbf{36.4}                 \\ \hline
\end{tabular}
\caption{PSNR performance of different methods in both linear and sRGB spaces. The results of the methods that cannot perform denoising are not included for the noisy scenario. The color space in the brackets indicates the particular color space of the employed training dataset. We exclude PSNR scores of methods who are pure demosaicking from the noisy columns.}
\label{tbl:msr}
\end{table}

\subsection{Demosaicking Non-Bayer CFA Data}
Finally, we explore the applicability of our approach to other Non-Bayer CFA, namely the Fuji-XTrans used by all modern Fuji digital cameras. Obviously, methods capable of demosaicking images from any camera sensor are preferred for practical applications. The images contained on MSR are also provided with the Fuji Xtrans CFA  but only on the noise-free case. Using this available data, we trained our method without initialization, and our results are provided in Table.~\ref{tbl:xtrans}. Clearly, our method is capable of outperforming the state of the art by a 0.7 dB margin for $K=20$ while being trained with only 200 images with flipping augmentations. At the same time, our network is capable of outperforming previous approaches even when employing only $K=8$ iterations and using as input the mosaicked image without any proper initialization.  The lack of proper initialization constrains the meaningful minimum number of iterations to be at least $K=8$ since fewer iterations are not enough to achieve satisfactory performance.
\begin{table}[htbp]
\centering
\begin{tabular}{lcc}
\hline \hline

                                                                    & \multicolumn{2}{c}{noise-free} \\
\multicolumn{1}{l}{}                                 & linear        & sRGB          \\
\multicolumn{1}{l}{\textbf{Trained on MSR Dataset:}} &               &               \\ \hline
Khashabi \cite{khasabi2014}                        & 36.9          & 30.6          \\
Klatzer \cite{klatzer2016}                         & 39.6          & 33.1          \\
Gharbi \cite{Gharbi:2016:DJD:2980179.2982399}      & 39.7          & 33.2   \\ 
Kokkinos \cite{kokkinos.2018}      & 39.9          & 33.7   \\ 
 $\text{MMNet}_{2}$  (ours)                                                 & 37.5 & 31.8 \\
$\text{MMNet}_{8}$  (ours)                                                 & 40.2 & 34.0 \\
$\text{MMNet}_{20}$  (ours)                                                 & \textbf{40.6} & \textbf{34.4} \\ \hline      
\end{tabular}
\caption{Evaluation on noise-free linear data with the non-Bayer mosaick pattern Fuji XTrans.}
\label{tbl:xtrans}
\end{table}

\subsection{ Running Time}
To evaluate the time complexity of the most recent state-of-the art methods, we estimated their execution time for an 1 Mpixel image using the source code that has been made publicly available by the respective authors. The benchmarks were run on a machine with a single NVIDIA GTX 1080 Ti GPU, an Intel i7-6850K CPU and 32 GB of RAM. Our method is capable of demosaicking at $2\times$ faster than Gharbi~\cite{Gharbi:2016:DJD:2980179.2982399} and $22\times$ than Henz~\cite{Henz.2018}, when the number of total iterations is kept small. However, as we increase the number of iterations the computation cost increases almost linearly. For example, as presented in Fig.~\ref{fig:heatmap} doubling the number of iterations from 10 to 20, while it yields a quantitative improvement of 0.2 dB in the reconstruction quality, it also requires twice the execution time. Therefore in practice, there is a trade-off between fast execution time and reconstruction quality, which needs to be taken into account by the users.

\begin{table}[h]
\centering
\begin{tabular}{|c|c|c|c|c|c|}
\hline
\multicolumn{6}{|c|}{Computation Time (sec/Mpixel)} \\ \hline
\multicolumn{2}{|c|}{Gharbi et al.} & \multirow{2}{*}{Henz et al.} & \multicolumn{3}{c|}{MMNet} \\ \cline{1-2} \cline{4-6} 
Bayer & Xtrans &  & 2 & 10 & 20 \\ \hline
0.37 & 0.87 & 2.88 & 0.13 & 1.11 & 2.34 \\ \hline
\end{tabular}
\caption{Running time comparison for the reconstruction of an one Mpixel image. The presented run times are the average of 10 runs on a NVIDIA GTX 1080Ti GPU.}
\label{tbl:computational_cost}
\end{table}

\section{Discussion}
There is a close connection between our proposed Algorithm \ref{alg:the_alg} and some instances of the proximal gradient descent algorithm (PGD), such as the Iterative Shrinkage Thresholding Algorithm (ISTA)~\cite{daubechies2004iterative} and its accelerated version (FISTA)\cite{fista.2009}. In fact, the denoising step \eqref{eq:upperboundeq} of our approach is equivalent to computing the proximal operator. However, between the two algorithms above and our approach, there are two differences that we would like to highlight. Firstly, a distinctive difference between our CNN based approach and the proximal operator is the fact that our denoiser can only approximate the solution, so in a sense, it is an inexact proximal solution. Thus our proposed algorithm acts as an Inexact Proximal Gradient Descent (IPGD)\cite{gu.2016}. Second, ISTA and FISTA require the exact form of the employed regularizer, such as Total Variation~\cite{rudin1992nonlinear} or Hessian Schatten-norm regularization~\cite{lefkimmiatis.2012,lefkimmiatis.2013}. In contrast, our method implicitly learns the regularizer from available data as a part of the proximal approximation. Of course, there is no straightforward way to derive the type of regularization that our deep learning denoiser has learned during training. However, this does not introduce any problems since there is no need to know either the regularizer nor the exact proximal solution. The reason for that is that as it is shown in~\cite{gu.2016}, even in the non-convex cases like ours, an inexact proximal solution can converge in the same convergence rate as the original PGD algorithms, provided that certain assumptions apply.

\section{Conclusion}
In this work, we presented a novel deep learning system that produces high-quality images from raw sensor data. Our demosaick network yields superior results both quantitative and qualitative compared to the current state-of-the-art solutions. Meanwhile, our approach is able to generalize well even when trained on small datasets and the number of our network parameters is kept low compared to other competing networks. Finally, we introduced an efficient way to train iterative networks that involve an arbitrary number of iterations. We hope that our training strategy, which is not specific to our network architecture, will pave the way to successful training of learning-based iterative approaches that target other ≈ image restoration tasks.

\ifCLASSOPTIONcaptionsoff
  \newpage
\fi

\bibliographystyle{IEEEtran}
\bibliography{egbib}

\end{document}